  \documentclass[sigconf,natbib]{acmart}
\usepackage{enumitem}
\usepackage{array}

\AtBeginDocument{%
  \providecommand\BibTeX{{%
    \normalfont B\kern-0.5em{\scshape i\kern-0.25em b}\kern-0.8em\TeX}}}

\setcopyright{rightsretained} 

\begin{document}
\copyrightyear{2023}
\acmConference[SIGIR'23]{The SIGIR '23 Workshop on Knowledge Discovery from Unstructured Data in Financial Services (KDF)}{July 23-27, 2023}{Taipei, Taiwan}
\acmDOI{}
\acmISBN{}
\title{Cross-Lingual NER for Financial Transaction Data in Low-Resource Languages}

\author{Sunisth Kumar}
\affiliation{%
  \institution{Genify}
  \city{Gurugram}
  \country{India}}
\email{ksunisth@gmail.com} 
\orcid{0000-0002-5290-1447}

\author{Davide Liu}
\affiliation{%
  \institution{Genify}
  \city{Beijing}
  \country{China}}
\email{davide@genify.ai}
\orcid{0000-0003-0380-404X}

\author{Alexandre Boulenger}
\affiliation{%
  \institution{Genify}
  \city{Abu Dhabi}
  \country{UAE}}
\email{alex@genify.ai}
\orcid{0000-0001-8120-0001}

\renewcommand{\shortauthors}{Kumar, et al.}

\begin{abstract}


We propose an efficient modeling framework for cross-lingual named entity recognition in semi-structured text data. Our approach relies on both knowledge distillation and consistency training. The modeling framework leverages knowledge from a large language model (XLMRoBERTa) pre-trained on the source language, with a student-teacher relationship (knowledge distillation). The student model incorporates unsupervised consistency training (with KL divergence loss) on the low-resource target language.

We employ two independent datasets of SMSs in English and Arabic, each carrying semi-structured banking transaction information, and focus on exhibiting the transfer of knowledge from English to Arabic. With access to only 30 labeled samples, our model can generalize the recognition of merchants, amounts, and other fields from English to Arabic. We show that our modeling approach, while efficient, performs best overall when compared to state-of-the-art approaches like DistilBERT pre-trained on the target language or a supervised model directly trained on labeled data in the target language.

Our experiments show that it is enough to learn to recognize entities in English to reach reasonable performance on a low-resource language in the presence of a few labeled samples of semi-structured data. The proposed framework has implications for developing multi-lingual applications, especially in geographies where digital endeavors rely on both English and one or more low-resource language(s), sometimes mixed with English or employed singly.
\end{abstract}





\maketitle

\section{Introduction}
Named Entity Recognition (NER) has become an essential task in Natural Language Processing (NLP) within the finance domain, given the exponential growth of digital content and the need to extract meaningful insights from financial texts. NER involves identifying and classifying named entities such as organizations, currencies, financial instruments, and monetary values, which is crucial for various NLP applications in finance, including sentiment analysis, risk assessment, and investment recommendation systems. However, developing an accurate and efficient cross-lingual NER model poses significant challenges, such as the lack of labeled data in low-resource languages and the difficulty of capturing cross-lingual variations.

Cross-lingual NER has significant implications on the industry, especially for companies that operate in numerous markets and need to analyze customer feedback, social media posts, and other types of unstructured data in multiple languages. Accurately identifying named entities in these languages can help companies extract valuable insights, identify trends, and make informed business decisions. However, developing accurate cross-lingual NER models requires a significant amount of resources, including labeled data, expertise in multiple languages, and computational resources. Therefore, there is a need for efficient and effective cross-lingual NER models that can transfer knowledge from high-resource languages to low-resource languages.

In this paper, we propose a novel framework to improve cross-lingual named entity recognition in semi-structured text data. The proposed framework leverages knowledge distillation to transfer knowledge from a teacher model pre-trained in a high-resource language (English) to a smaller student model. Next, we employ consistency training to fine-tune the student model to Arabic, a low-resource language \cite{10.1007/978-3-030-58586-0_20}. Our experiments focus on recognizing entities in semi-structured SMSs that carry banking transaction information in English and Arabic, with access to only a few labeled examples in the target language.

The proposed model shows remarkable cross-lingual learning ability and outperforms state-of-the-art models directly trained on the target language. The main contributions of our work are as follows:
\begin{itemize}[topsep=5pt] 
    \item We leverage knowledge distillation and consistency training to enhance cross-lingual NER in semi-structured text.
     \item This approach is efficient and effective, requiring only a few labeled examples in the target language. 
    \item The resulting model outperforms other approaches, including DistilBERT pre-trained on the target language or a supervised model trained directly on labeled data in the target language, especially for multi-lingual entity recognition.
\end{itemize}

\section{Related Works}
The problem of cross-lingual Named Entity Recognition (NER) in low-resource languages presents unique challenges and has gathered significant attention in recent research. In recent years, a lot of focus has been on using transfer learning-based methods to address this problem. These methods leverage pre-trained models such as BERT and XLM-RoBERTa to improve the performance of NER models.

Knowledge Distillation has gained attention as a transfer learning technique for cross-lingual NER \cite{ma-etal-2022-wider, wu-etal-2020-single}. By transferring the knowledge from a pre-trained model (teacher) to a smaller model (student), knowledge distillation enables the student model to benefit from the rich representations learned by the teacher model. This method allows for the effective utilization of large-scale pretraining while adapting to the specific cross-lingual NER task.

In addition to the knowledge distillation, we propose incorporating consistency training into the cross-lingual NER framework. Consistency training \cite{zhou-etal-2022-conner, wang-henao-2021-unsupervised} uses an unsupervised loss function to measure the consistency of the model's predictions on the same input with small random perturbations. By encouraging the model to produce consistent outputs under perturbations, the generalization capability of the model can be enhanced, leading to improved cross-lingual NER performance.



\section{Methods}
In this section, we present the methodology employed for the problem of cross-lingual NER for semi-structured financial text data in low-resource languages. This section is structured into three sub-sections: Problem Formulation, Model Architecture, and Training.

\subsection{Problem Formulation}
We formulate the task of cross-lingual NER as follows:

Let an input text sequence \(X = \{x_1,x_2,...,x_n\}\), where $n$ is the length of the sequence. Each token $x_i$ is associated with a label $y_i$, representing its NER tag. The set of possible NER tags is denoted as \(Y = \{y_1,y_2,...,y_k\}\), where $k$ is the total number of entity types.

Given a set of labeled data \(D = \{(X_1,Y_1),(X_2,Y_2),...,(X_m,Y_m)\}\), where each \((X_i,Y_i)\) pair represents an input text sequence and its corresponding NER tag, the objective is to learn a model $M$ that can accurately predict the NER tag $Y_i$ for an unseen input sequence $X_i$ in different languages. In this paper, we are using the Arabic language as a low-resource language.

The cross-lingual aspect of the problem arises from the scarcity of labeled data in low-resource languages. Therefore, the model $M$ should be capable of transferring knowledge from high-resource languages such as English, to low-resource languages, such as Arabic, to improve the performance of NER in those languages.

\subsection{Model Architecture}
To address these challenges of the cross-lingual NER task, we propose a novel framework based on knowledge distillation and consistency training. The model architecture is designed to leverage the benefits of both student-teacher knowledge distillation and consistency training.

\begin{figure}[htbp]
  \centering
  \includegraphics[width=0.47\textwidth]{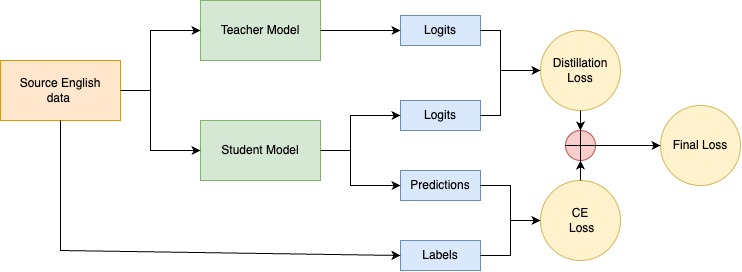}
  \caption{Overview of the student-teacher training framework (KD) with knowledge distillation and cross-entropy loss training on English data.}
  \label{fig:kd}
\end{figure}

\begin{figure}[htbp]
  \centering
  \includegraphics[width=0.47\textwidth]{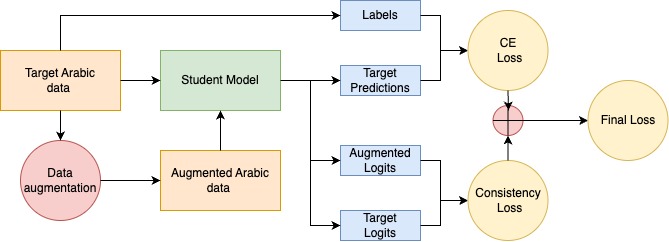}
  \caption{Overview of the knowledge distillation and consistency training framework (KD+CT) for training on Arabic data with consistency-training and cross-entropy loss.}
  \label{fig:ct}
\end{figure}

The overall architecture consists of two components, namely (1) knowledge distillation with supervised cross-entropy loss and (2) consistency training.

\subsubsection{Teacher Model}
The teacher model $T$ is a pretrained XLM-RoBERTa model \cite{conneau-etal-2020-unsupervised}, fine-tuned on the source language (English) dataset. It serves as the source of knowledge transfer and provides soft target distributions for the student model during training. The teacher model takes input tokens $X$ and produces token-level predictions \(P^T = \{p^T_1,p^T_2,...,p^T_n\}\), where $p^T_i$ represents the probability distribution over the NER tags for the token $x_i$.

\subsubsection{Student Model}
The student model $S$ is a DistilBERT model \cite{Sanh2019DistilBERTAD}. It consists of a multi-layer transformer encoder, similar to the teacher model but with fewer layers and smaller hidden dimensions. The student model takes input tokens $X$ and produces token-level predictions \(P^S = \{p^S_1,p^S_2,...,p^S_n\}\), where $p^S_i$ represents the probability distribution over the NER tags for the token $x_i$.

\subsubsection{Knowledge Distillation}
We use knowledge distillation to transfer knowledge from the teacher model to the student model, to reduce the model size. The distillation loss combined with supervised cross-entropy loss (i.e., \(\mathcal{L}_{\text {CE}})\) is defined as: 
\begin{equation}
\mathcal{L}_{\text {distill}}=\alpha\mathcal{L}_{\text {CE}} + (1-\alpha)\mathrm{KL}\left(P^T \| P^S\right)
\end{equation}
where \(\alpha\) is the weight coefficient, and $P^T$ and $P^S$ represent the softened probability distributions obtained by applying the softmax function to the logits of the teacher model and the student model, respectively.

\subsubsection{Consistency Training}
After the knowledge distillation training, the student model is fine-tuned in the target language (Arabic) using consistency training. Consistency training encourages the model to produce consistent predictions when given different perturbations of the same input. We use a combination of supervised cross-entropy loss (i.e., \(\mathcal{L}_{\text {CE}})\) and the unsupervised KL divergence as the consistency loss, comparing the predictions of the augmented data and the original data:
\begin{equation} \label{eq:2}
\mathcal{L}_{\text {consistency}}=\alpha\mathcal{L}_{\text {CE}} + (1-\alpha)\mathrm{KL}\left(P^{\text{augmented}} \| P^{\text{original}}\right)
\end{equation}
where \(\alpha\) is the weight coefficient, and $P^{\text{augmented}}$ and $P^{\text{original}}$ represent the softmax probabilities obtained from the augmented data and the original data, respectively.

During consistency training, we generate augmented versions of the target language data \cite{xie2019unsupervised} using back translation, RandAugment, and TF-IDF word replacement. These augmented examples are used to compute the unsupervised consistency loss and update the student model parameters accordingly.

% ...

\section{Experiments}

\subsection{Dataset}

\begin{table}[h]
\centering
\begin{tabular}{|c|c|c|} 
\hline
\textbf{Entity} & \textbf{English Dataset} & \textbf{Arabic Dataset} \\ \hline
amount & 3511 & 73 \\ \hline
supplier & 2968 & 29 \\ \hline
currency & 2490 & 34 \\ \hline
number & 2465 & 34 \\ \hline
full-date & 2234 & - \\ \hline
card-number & 1951 & 7 \\ \hline
full-time & 1938 & - \\ \hline
merchant & 1133 & 7 \\ \hline
balance & 494 & 8 \\ \hline
time & 135 & 8 \\ \hline
month & 99 & 2 \\ \hline
date & 10 & 43 \\ \hline
\textbf{Total Entities} & \textbf{19428} & \textbf{221} \\ \hline
\end{tabular}
\caption{Unique Named Entities in English and Arabic Datasets. Each row represents a specific named entity, and the corresponding columns indicate the count of occurrences for that entity in each dataset.}
\label{tab:entities}
\end{table}

We conduct our experiments on a financial transactions dataset consisting of semi-structured SMS data in English and Arabic. The dataset is sourced from Egypt. The English language dataset consists of 1730 sentences along with associated annotated NER tags. The Arabic language dataset consists of 30 sentences. Both language datasets were preprocessed to hide sensitive information and converted to the standard IOB format for NER before training. The detailed distribution of unique named entities in these datasets can be found in Table \ref{tab:entities}. The Arabic language dataset is used unlabeled for the consistency loss and labeled for the supervised loss. The augmented dataset is generated from this original dataset in the Arabic language.

\subsection{Experimental Setup}

\begin{figure*}[htbp]
  \centering
  \includegraphics[width=\textwidth]{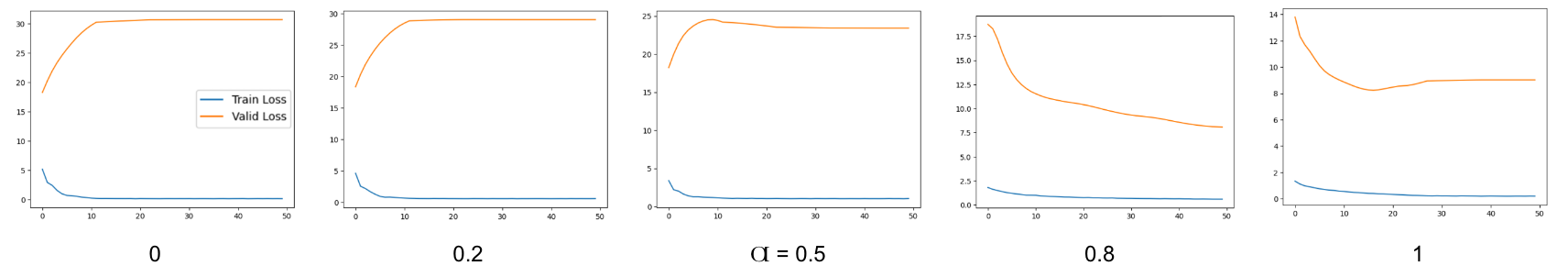}
  \caption{The training and validation loss of the KD+CT model (our model) over different values of $\alpha$ in $[0, 0.2, 0.5, 0.8, 1]$ when training it on Arabic data. The loss value and the number of epochs are on the $y$-axis and $x$-axis, respectively. The results indicate that when setting the value of $\alpha$ to $0$ and $0.2$, the model exhibits overfitting behavior on the validation data, as evidenced by an increase in validation loss while the training loss continues to decrease. For $\alpha$ equal to $0.5$ and $1$, overfitting is still present but not so severe as it was for smaller $\alpha$. Finally, we empirically found that $\alpha = 0.8$ shows the most desirable learning behavior for validation loss, which almost linearly decreases for the duration of the training.}
  \label{fig:loss}
\end{figure*}

We implement our NER model using the Transformers library and the BERT-based architecture for the teacher and student models. We use AdamW optimizer \cite{loshchilov2018decoupled} with a learning rate \(l_r = 2e-5\). We use a batch size of 28 and train the NER model for $20$ epochs.

We run experiments on $\alpha$ over the range of ${0,0.2,0.5,0.8,1}$, (as shown in Figure \ref{fig:loss}). We set $\alpha = 0.8$ based on the best performance. At $\alpha = 0$ (i.e., only unsupervised consistency training loss), the NER model does not learn, and the validation loss increases. At $\alpha = 1$ (i.e., only supervised cross-entropy loss), we observe overfitting on the limited target language data (Arabic), and the validation loss starts to increase after going down. However, at $\alpha = 0.8$ (combination of supervised and unsupervised losses), the NER model gives the best performance on the cross-lingual NER task for low-resource language.

\subsection{Performance Comparison}
To evaluate the performance of our proposed cross-lingual NER model, we compare it with that of several baseline models and existing state-of-the-art approaches. The baselines include:

\begin{enumerate}
    \item Teacher Model: A pretrained large language model (XLM-RoBERTa) fine-tuned on the English language dataset.
    \item Student Model: A DistilBERT-based student model trained using knowledge distillation from the teacher model.
    \item Naive Benchmark Model: A pretrained DistilBERT model fine-tuned on the target language (Arabic) dataset.
\end{enumerate}

We report the performance comparison in terms of F1 score and accuracy for NER on both the source (English) and the target (Arabic) datasets. 

\subsection{Results and Analysis}

\begin{table}[h]
\centering
\begin{tabular}{|c|c|c|c|c|}
\hline
\textbf{} & \multicolumn{2}{|c|}{\textbf{English}} & \multicolumn{2}{|c|}{\textbf{Arabic}} \\ \hline
\textbf{Model} & \textbf{F1} & \textbf{Acc} & \textbf{F1} & \textbf{Acc} \\ \hline
Teacher & 0.9887 & 0.9888 & 0.5929 & 0.6543 \\ \hline
Student (only KD) & \textbf{0.9957} & \textbf{0.9957} & 0.5693 & 0.6852 \\ \hline
Student (KD+CT) & 0.9768 & 0.9782 & \textbf{0.6540} & \textbf{0.7407} \\ \hline
DistilBERT & 0.6263 & 0.7377 & 0.6065 & 0.7099 \\ \hline
\end{tabular}
\caption{Comparison of the NER performance of the models on English and Arabic datasets. The accuracies and F1 scores are shown for both English and Arabic datasets.}
\label{tab:results}
\end{table}

We compare the performance of our NER model with the Teacher model, the Student model, and the Naive Benchmark model on both the source (English) and the target (Arabic) datasets.

On the English dataset, our model achieves an F1 score of $0.9768$.
Although the Teacher and Student models exhibit higher F1 scores, we note that our model achieves comparable performance while being smaller than the Teacher model, with an F1 score of $0.9887$. On the Arabic dataset, our model significantly outperforms the Teacher and the Student models, reaching an F1 score of $0.6540$ and an accuracy of $0.7407$. Furthermore, our model performs better than the Naive Benchmark model having an F1 score of $0.6065$. 

These results show that our model achieves competitive performance on both the English (source) and the Arabic (target) datasets. Despite its smaller size and the limited data available in the target language, our model demonstrates remarkable cross-lingual generalization capabilities. It effectively leverages the knowledge distilled from the Teacher model and further enhances its performance through consistency training on the limited target language data.

The overall superior performance of our model can be attributed to its ability to capture and transfer the underlying patterns learned by the Teacher model, leveraging the knowledge distilled during the training process. By incorporating consistency training, our model achieves more robust predictions by ensuring consistency across augmented versions of the input sequences. This training mechanism enhances the model’s ability to adapt to cross-lingual contexts and improve performance. The successful combination of knowledge distillation and consistency training contributes to the model’s superior performance in capturing both the general patterns and specific language characteristics required for effective cross-lingual named entity recognition.

Overall, our proposed cross-lingual NER model emerges as a promising approach for low-resource languages. Its ability to achieve competitive performance with a smaller model size makes it a practical and efficient solution for real-world applications.


\section{Conclusion}
In this paper, we introduced a novel framework that uses knowledge distillation and consistency training to enhance cross-lingual named entity recognition in semi-structured text data. Knowledge is transferred from a teacher model pre-trained in English to a smaller student model, which is then fine-tuned for Arabic. We validated the performance of our model (KD+CT) on semi-structured banking transaction data in both English and Arabic, showing competitive performance on both datasets.

Our experiments highlight the potential of our modeling approach to combine knowledge distillation with consistency training and to address the significant challenges of developing accurate and efficient cross-lingual NER models in low-resource languages. 

Our model significantly outperforms the naive benchmark, the student, and the teacher models in entity recognition on the target language dataset (Arabic) and achieves performance comparable to the larger teacher model while being smaller in size on the source language dataset (English). This demonstrates the remarkable cross-lingual generalization capabilities of our model.

Our model demonstrates remarkable performance in entity recognition on the target language dataset (Arabic), outperforming the naive benchmark, the student, and the teacher models. Additionally, our model achieves comparable performance to the larger teacher model on the source language dataset (English) while maintaining a smaller size. These results highlight the exceptional cross-lingual generalization capabilities of our model.


We believe that our proposed cross-lingual NER model can contribute to the development of multi-lingual applications and enable companies to extract insights, identify trends, and make informed business decisions in multiple languages. We hope our work inspires further research in this field and facilitates the development of efficient and effective cross-lingual NER models for low-resource languages and beyond.

\bibliographystyle{ACM-Reference-Format}
\bibliography{refs}

\end{document}